\documentclass[11pt]{article}

\usepackage{acl}

\usepackage{times}
\usepackage{latexsym}
\usepackage[T1]{fontenc}
\usepackage[utf8]{inputenc}
\usepackage{microtype}

\usepackage{amsmath,amssymb,amsfonts}
\usepackage{booktabs}
\usepackage{array}
\usepackage{multirow}
\usepackage{graphicx}
\usepackage[ruled,lined,linesnumbered]{algorithm2e}
\SetKwInput{KwInput}{Input}
\SetKwComment{tcp}{// }{}
\DontPrintSemicolon
\usepackage{pifont}

\newcommand{\ours}{SCOPE}
\newcommand{\best}[1]{\textbf{#1}}

\title{Revise, Don't Freeze: Sampler-Matched Training\\for Self-Correcting Masked Diffusion Language Models}
\author{
  \textbf{Longxuan Yu\textsuperscript{1}\thanks{Equal contribution.}},
  \textbf{Shaorong Zhang\textsuperscript{1}\footnotemark[1]},
  \textbf{Yu Fu\textsuperscript{1}\footnotemark[1]},
  \textbf{Hui Liu\textsuperscript{2}},
\\
  \textbf{Yue Dong\textsuperscript{1}},
  \textbf{Greg Ver Steeg\textsuperscript{1}\thanks{Corresponding author: \texttt{gregoryv@ucr.edu}}}
\\
\\
  \textsuperscript{1}University of California, Riverside,
  \textsuperscript{2}Microsoft
}

\begin{document}
\maketitle

\begin{abstract}
Masked diffusion language models (MDLMs) re-predict every position at each denoising step, but standard samplers commit tokens once revealed, leaving this revision capability unused. Existing approaches either add heuristic or learned mechanisms to revise committed tokens, or remask them back to [MASK] before re-predicting; a principled sampler that directly revises visible tokens without auxiliary modules remains underexplored. We introduce \textbf{D3IM}, a parameter-free sampler derived as a corrector-style reverse update that permits direct visible-to-visible revision without additional modules or auxiliary passes. D3IM also reveals a model-side obstacle we term \emph{preservation bias}: the model tends to reproduce its own wrong committed tokens rather than correct them. We address this with \textbf{\ours{}} (Self-Conditioned On Prediction Errors), a lightweight post-training procedure that simulates D3IM's sampling process. On LLaDA-8B at 64 denoising steps, \ours{}+D3IM improves over the original LLaDA-8B with standard unmasking by +13.0 on GSM8K (68.3\%), +4.8 on MATH-500 (23.6\%), +15.3 on HumanEval (29.3\%), and +10.4 on MBPP (30.8\%), with gains that increase as more denoising steps are used on math and HumanEval.

\end{abstract}

\section{Introduction}

Masked diffusion language models (MDLMs) generate text by iteratively denoising masked sequences, re-predicting every position at each step~\cite{d3pm,sedd,mdlm,llada}. This all-position prediction naturally creates a revision capability: even after a token becomes visible, the denoiser can assign a different prediction to that position at a later step. However, standard unmasking-based MDLM decoding freezes each token once it is committed, updating only masked positions and discarding later predictions at visible positions. Early mistakes therefore become permanent, a cost especially severe on reasoning-heavy tasks such as math and code generation, where a single wrong digit or operator can constrain subsequent predictions~\cite{early_decisions}.

\begin{figure*}[t]
\centering
\includegraphics[width=\textwidth]{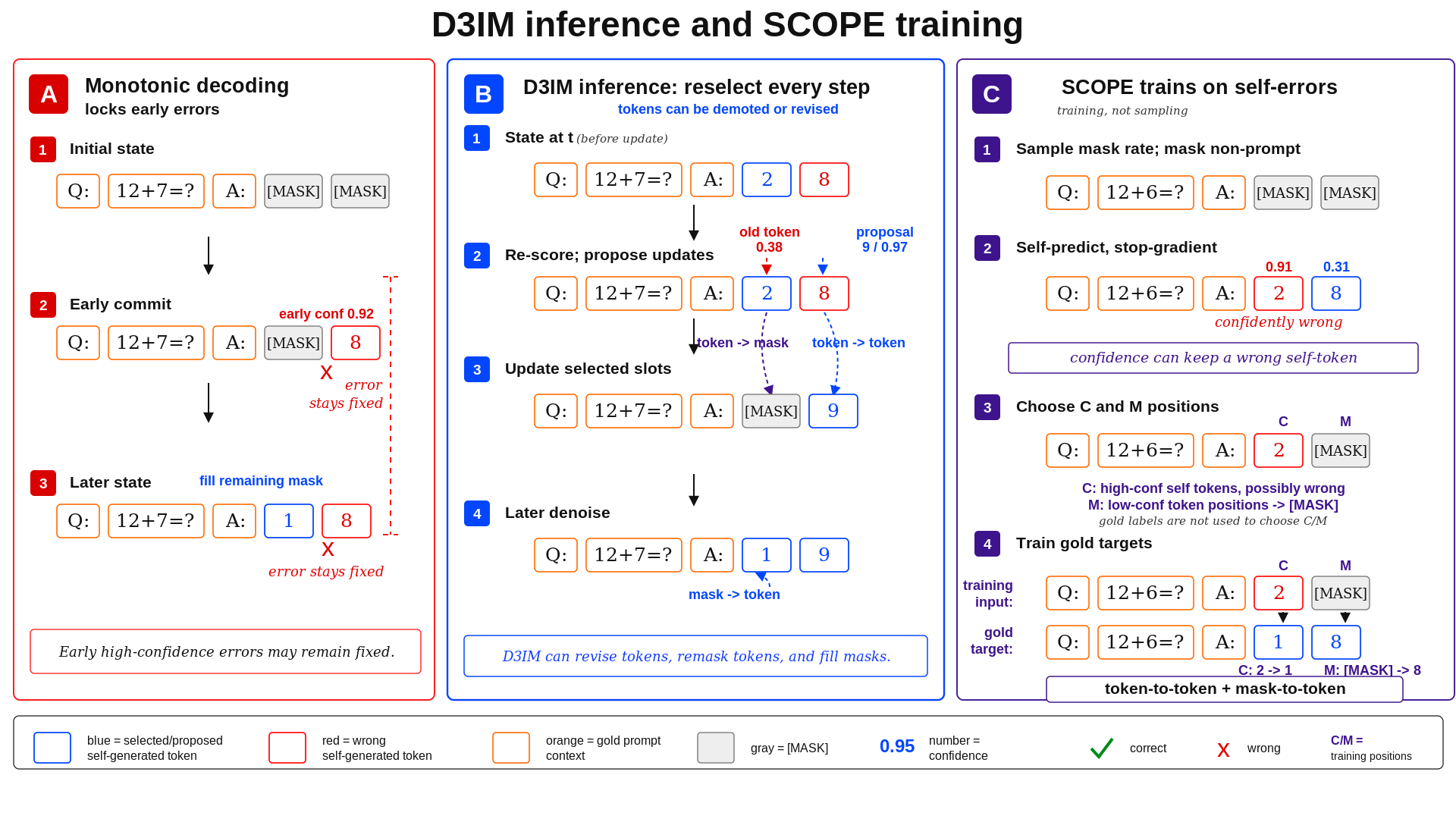}
\caption{Overview of D3IM inference and \ours{} training. \textbf{(A)}~Standard monotonic unmasking freezes tokens once committed; early high-confidence errors remain fixed. \textbf{(B)}~D3IM re-predicts all positions at each step and keeps only the top-$K$ by confidence. A low-confidence token is demoted to [MASK] (token$\to$mask); a high-confidence new prediction directly overwrites the old token (token$\to$token). \textbf{(C)}~\ours{} training simulates this process: the model self-predicts on masked input (step~2), and the top-$\rho$ predictions by confidence are committed as set $C$ while the rest remain masked as set $M$ (step~3). The loss is computed on both $C$ and $M$ positions against ground-truth targets (step~4), training token-to-token and mask-to-token correction jointly.}
\label{fig:overview}
\end{figure*}

Existing revision approaches fall short in two complementary ways. The closest principled remask sampler directly applicable to pretrained MDLMs, ReMDM~\cite{remdm}, is derived under the absorbing assumption and can only revise a committed token by first returning it to [MASK]. Practical revision methods bypass this remasking route but add heuristic confidence thresholds~\cite{llada21,t2m}, verification passes~\cite{wino,tolerator}, or learned modules~\cite{remedi} on top of the base denoising loop. The denoiser, however, already re-predicts every position at every step, including committed ones. A direct visible-to-visible revision sampler that exploits these predictions without remasking or auxiliary modules remains largely unexplored.

We close both gaps with \textbf{D3IM} (\S\ref{sec:d3im}), a direct-revision sampler that re-predicts each intermediate sequence from current predictions rather than editing the previous state. \emph{Theoretically}, we derive D3IM as a corrector-style reverse update for pretrained MDLMs like LLaDA, in which visible tokens may transition directly to new token values rather than being copied or first returned to [MASK] (Appendix~\ref{sec:appendix_theory}). \emph{Practically}, D3IM is parameter-free and reuses the standard denoising forward pass without extra modules or auxiliary passes, under the same linear top-$K$ schedule as standard unmasking: at each step, it re-predicts all positions and retains the top-$K$ by current confidence, so a previously committed token survives only if the model continues to support it.

D3IM provides a mechanism for revision. However, directly applying it to a base MDLM trained only with the standard mask-only objective degrades performance. We trace this to \emph{preservation bias}: the model was never exposed to its own errors during training, and it overwhelmingly reproduces wrong committed tokens rather than correcting them (\S\ref{sec:bias_resolution}). To address this, we propose \textbf{\ours{}} (Self-Conditioned On Prediction Errors, \S\ref{sec:scope_training}), a lightweight post-training procedure that simulates D3IM's sampling process, exposing the model to its own high-confidence errors and training it to recover the ground truth. With \ours{}+D3IM, LLaDA-8B at 64 steps improves over the original LLaDA-8B with standard unmasking by +13.0 on GSM8K, +4.8 on MATH-500, +15.3 on HumanEval, and +10.4 on MBPP, with gains that increase as more denoising steps are used on math and HumanEval.

Our contributions:
\begin{enumerate}
\item We propose \textbf{D3IM}, a parameter-free direct-revision sampler for pretrained MDLMs. D3IM is derived as a corrector-style reverse update in which visible tokens may transition directly to new values rather than being copied or returned to [MASK] (\S\ref{sec:d3im}; Appendix~\ref{sec:appendix_theory}).
\item We identify \emph{preservation bias} as the model-side obstacle to revision: trained on ground-truth context, the model reproduces its own wrong predictions even when the training objective does not require it (\S\ref{sec:bias_resolution}).
\item We show that a lightweight post-training procedure (\textbf{\ours{}}) that simulates D3IM's sampling process is sufficient to substantially reduce preservation bias and make D3IM effective, without complex multi-objective training or architectural changes (\S\ref{sec:scope_training}).
\end{enumerate}


\section{Preliminaries}
\label{sec:background}

\paragraph{Absorbing MDMs.}
Classical masked diffusion models generate by gradually converting mask tokens into visible tokens~\cite{d3pm,sedd,mdlm}. In the absorbing formulation, once a coordinate becomes visible, the clean-token posterior collapses to the current token,
\begin{equation}
p(X_0^i \mid X_t, X_t^i \neq m) = \delta_{X_t^i}.
\end{equation}
Thus, a visible token is no longer a hypothesis to be re-evaluated; it is treated as the clean token itself. This makes monotonic denoising stable, but it also makes committed-token correction impossible: if an early token is wrong, later reverse steps have no mechanism to replace it. We provide the formal derivation in Appendix~\ref{sec:appendix_formal_background}.

\paragraph{LLaDA and behavioral absorption.}
LLaDA-style MDLMs weaken this constraint at the training-objective level. Their loss supervises only masked positions,
\begin{equation}
\mathcal{L}_{\mathrm{LLaDA}}
=
-\sum_{i \in \mathcal{M}_t}
\log p_\theta(x_i^* \mid \mathbf{x}_t),
\end{equation}
and imposes no loss on already visible tokens~\cite{llada}. The model is therefore not explicitly trained to satisfy the absorbing identity, and can in principle assign a different prediction to a visible position. However, this does not mean that absorption is solved. Standard LLaDA inference still freezes committed tokens, and even when this freezing is removed, the model itself tends to preserve its own visible predictions. In our wrong-commit stress test (\S\ref{sec:bias_resolution}), LLaDA-8B-Instruct preserves its own wrong visible token 70.1\% of the time and recovers the ground-truth token only 9.2\%. We call this \emph{preservation bias}: LLaDA-style training removes the formal absorbing constraint, but absorption reappears as a learned behavior under self-generated context.

\paragraph{Existing revision methods.}
ReMDM~\cite{remdm} provides a principled way to introduce remasking into absorbing-state MDMs by deriving a modified backward remasking process. This is well matched to the classical setting where visible tokens are locked by the absorbing posterior. More recent revokable decoding methods~\cite{llada21,t2m,wino,tolerator,remedi} depart from this derivation and instead use confidence thresholds, draft-and-verify, or targeted token-to-mask heuristics. These methods are effective, but they do not ask the specific question raised by LLaDA-style training: if visible positions are not constrained to copy themselves, can every position be re-predicted and globally reselected by the model's current confidence?

Self-correcting inference for LLaDA-style MDLMs therefore requires solving two problems. First, the sampler must address \emph{sampler-level absorption}: previously committed tokens should not be automatically preserved; each token must be re-evaluated at each step. Second, the model must address \emph{behavioral absorption}: under self-generated context, wrong visible predictions should be revokable rather than reproduced. D3IM sampling addresses the first (\S\ref{sec:d3im}); \ours{} training addresses the second (\S\ref{sec:scope_training}).

\section{Method}
\label{sec:method}

Our method pairs a sampler with a training procedure (Figure~\ref{fig:overview}). D3IM changes inference: instead of updating an existing generated sequence, it reconstructs each intermediate state from the model's current predictions. \ours{} changes training: it exposes the model to the self-generated visible errors that D3IM creates, so the confidence ranking used by D3IM becomes reliable.

\subsection{D3IM: Clean-Slate Remasking}
\label{sec:d3im}

D3IM is a parameter-free sampler designed for LLaDA-style visible-position predictions. Unlike standard monotonic unmasking, D3IM does not copy a token from the previous state simply because it was committed earlier. At every denoising step, the model predicts all active positions, ranks them by confidence, and keeps only the top-$K$ current predictions. All other positions are reset to [MASK].

Concretely, at step $t$, the model produces a candidate token and confidence score for each position. D3IM selects the $K(t)=\lfloor Lt/T \rfloor$ positions with the highest confidence, fills those positions with the model's current predictions, and masks the rest. Since the candidate is recomputed from the current logits, a previously committed token can either survive, be demoted to [MASK], or be overwritten by a new prediction. Thus, D3IM enables both token-to-mask demotion and token-to-token revision using only the confidence ranking already used by MDLM decoding.

This clean-slate rule means the next state is determined by the current predictions rather than by copying previously visible tokens. In the corrector sampling framework for discrete diffusion~\cite{dfm}, each reverse step is a mixture of three operations weighted by $a_t$, $b_t$, and $c_t$ ($a_t + b_t + c_t = 1$): predict a new token from the model ($a_t$), remask to [MASK] ($b_t$), or preserve the current token ($c_t$). Standard absorbing samplers set $c_t$ high so that committed tokens are frozen. D3IM sets $c_t = 0$: no weight is placed on preserving the current token, and the masking schedule is maintained entirely through $a_t$ and $b_t$. This is a valid corrector update as long as the noise-conservation constraint is satisfied at each step (full derivation in Appendix~\ref{sec:appendix_theory}; pseudocode in Appendix~\ref{sec:algo_details}). As a result, visible-to-visible revision is naturally supported within the framework without routing through [MASK]. However, because every position is re-evaluated by confidence, D3IM's effectiveness depends on how well the model's confidence reflects actual correctness under self-generated context (\S\ref{sec:method_pairing}).

\subsection{\ours{}: Training on Self-Generated Error States}
\label{sec:scope_training}

\ours{} training exposes the model to the kind of intermediate states that D3IM sampling creates. The goal is not simply to improve masked-token prediction, but to train the model to treat visible self-generated tokens as hypotheses rather than fixed context.

For each self-conditioning step, we first mask a random subset of non-prompt positions. The model then performs a no-gradient forward pass on this partially masked input and samples predictions for the masked positions. We commit only the top-$\rho$ fraction of these sampled predictions by confidence, leaving the remaining masked positions as [MASK]. The resulting training input contains three types of context: ground-truth visible tokens, self-generated committed tokens, and remaining masks. Crucially, the committed tokens include high-confidence errors, which are exactly the states where preservation bias is most pronounced.

We then run a gradient-enabled forward pass on this constructed input and supervise recovery to the ground-truth sequence only on the committed and still-masked positions:
\[
\mathcal{L}_{\ours{}}
=
\frac{1}{|C \cup M|}
\sum_{i \in C \cup M}
\mathrm{CE}(f_\theta(x_{\mathrm{train}})_i, x_i^*),
\]
where $C$ is the set of committed self-predictions and $M$ is the set of remaining masked positions. Ground-truth context positions are excluded from the loss, since they are already correct and would dilute the correction signal.

In practice, we sample the mask rate from Beta(2,2), use temperature sampling to increase the number of self-prediction errors used for correction training, and set the commit rate to $\rho=0.3$. With probability 0.5, we replace the self-conditioning branch with standard MDLM training to preserve ordinary masked-token prediction ability. Full pseudocode and implementation details are provided in Appendix~\ref{sec:algo_details}.

\subsection{Why the Pairing Matters}
\label{sec:method_pairing}

D3IM sampling and \ours{} training solve different parts of the same problem. D3IM constructs each intermediate state from current predictions, so every visible token must be selected again by the current top-$K$ ranking. But this exposes preservation bias: the original model often assigns high confidence to its own wrong visible tokens. \ours{} changes the model's behavior under exactly these self-generated contexts. It does not introduce a new decoding rule; it makes D3IM's existing top-$K$ confidence ranking reliable. Although \ours{} changes only training, the benefit is sampler-specific: it aligns confidence with D3IM's clean-slate top-$K$ rule and does not automatically transfer to other revokable samplers (\S\ref{sec:sampler_specificity}).

\section{Experiments}
\label{sec:experiments}

\subsection{Setup}

\paragraph{Base model.} LLaDA-8B-Instruct~\cite{llada}.

\paragraph{Training.} LoRA ($r{=}64$, $\alpha{=}128$) on attention projections (q, k, v, o), 50.3M trainable parameters (0.6\% of 8B). 500 steps on FineWeb-Edu~\cite{fineweb_edu} (streaming), per-device batch size 1 with gradient accumulation 4, lr=$1.5{\times}10^{-4}$, constant with warmup. Single A100 GPU, ${\sim}$20 minutes. Self-conditioning probability 0.5, commit rate $\rho{=}0.3$, temperature $\tau{=}1.5$.

\paragraph{Inference.} D3IM remasking (Algorithm~\ref{alg:D3IM}) with $T \in \{64, 128, 256\}$ denoising steps, single-block semi-AR decoding, greedy argmax. Generation length is 256 (GSM8K, HumanEval, MBPP) or 512 (MATH-500). EOS/EOT tokens receive confidence $-\infty$ during non-final steps to prevent premature termination (ablated in Appendix~\ref{sec:eos_policy_ablation}).

\paragraph{Benchmarks.} GSM8K~\cite{gsm8k} (1319 problems, 8-shot CoT), MATH-500~\cite{math500} (500 problems, zero-shot, sympy-verified), HumanEval~\cite{humaneval} (164 problems, pass@1), and MBPP~\cite{mbpp} (500 problems, 3-shot pass@1 via \texttt{lm-evaluation-harness}). For each benchmark, prompt, generation length, decoding budget, and evaluation protocol are fixed across rows; accuracy differences in Table~\ref{tab:main} reflect only the model checkpoint and sampling strategy.

\paragraph{Baselines.} Original LLaDA-8B-Instruct (no additional training). For calibration analysis (\S\ref{sec:calibration}) we additionally compare against XDLM~\cite{adaptive_order} and standard MDM continued pretraining (MDM-CPT). For inference-only revokable samplers we evaluate ReMDM-conf, ReMDM-cap~\cite{remdm}, and Tolerator~\cite{tolerator} at their published default hyperparameters (Table~\ref{tab:inference_baselines}). For qualitative trajectory evaluation we use a Claude-Sonnet-4-6 judge to categorize individual correction events as useful, neutral, or harmful (\S\ref{sec:trajectory}). Additional ablations (temperature, commit-rate schedule, sampler specificity) are reported in the Appendix.
\begin{figure*}[t]
\centering
\begin{minipage}[t]{0.48\textwidth}
\centering
\textbf{(a) ECE by mask rate}\\[+0.1em]
\includegraphics[width=\linewidth]{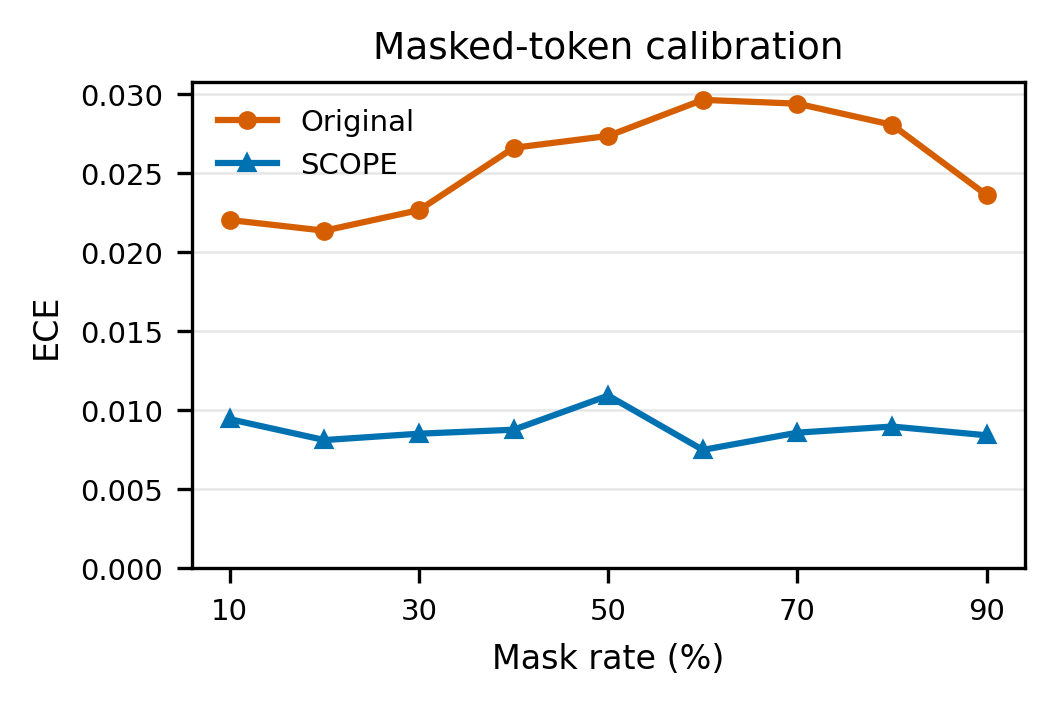}
\end{minipage}
\hfill
\begin{minipage}[t]{0.48\textwidth}
\centering
\textbf{(b) Wrong-commit stress}\\[+0.1em]
\includegraphics[width=\linewidth]{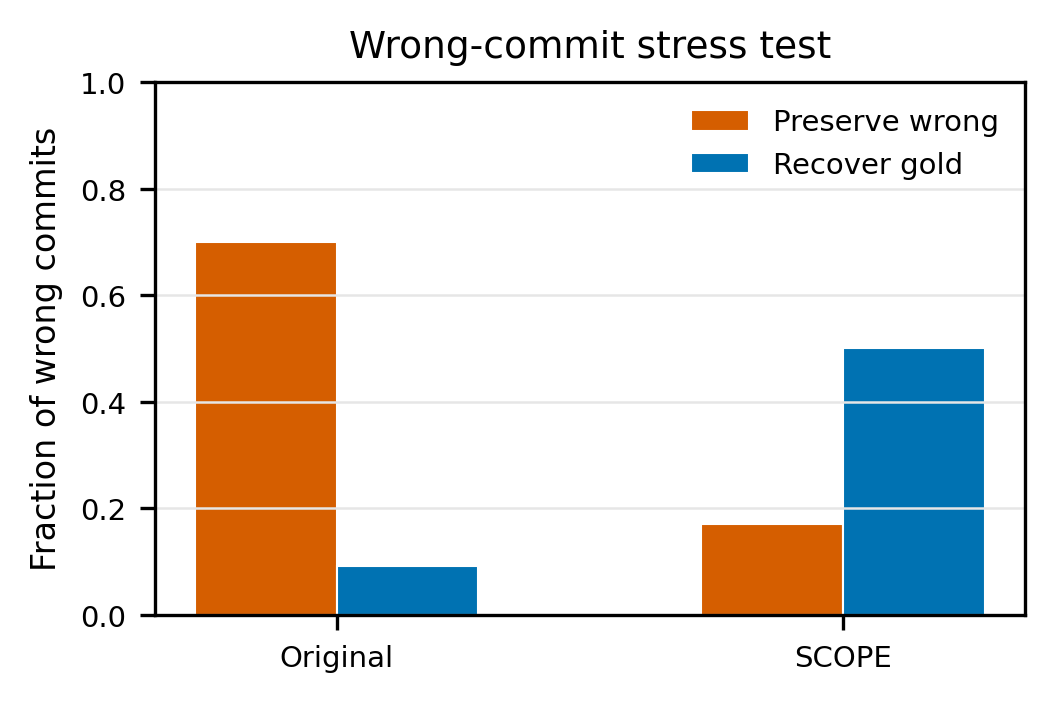}
\end{minipage}
\caption{Calibration and self-correction diagnostics. Left: masked-token ECE across mask rates on 900 held-out passages from ArXiv, CNN/DailyMail, and PubMed (300 each; 45K--407K masked tokens per point). Right: among positions where the self-committed token is wrong, Original usually preserves the wrong token, while \ours{} more often recovers the ground-truth token; residual cases switch to neither token. Together, these diagnostics show that \ours{} improves confidence behavior and makes erroneous commits more revokable under D3IM.}
\label{fig:calib_stress}
\end{figure*}

\subsection{Mechanism Diagnostics}
\label{sec:mechanism_diagnostics}

Before evaluating downstream tasks, we verify the mechanism that \ours{} is designed to change: confidence under masked context and the tendency to preserve self-generated mistakes.

\paragraph{Calibration.}
\label{sec:calibration}
Since D3IM uses confidence only through a top-$K$ ranking, we report both calibration-style ECE and ranking-style wrong-commit survival diagnostics. \ours{} reduces masked-token Expected Calibration Error relative to Original across all tested mask rates (Figure~\ref{fig:calib_stress}, left). In D3IM's iterative setting, a useful confidence ranking lets a token lose the top-$K$ competition and be re-predicted from updated context. This diagnostic suggests that ranking quality, not only raw masked-token accuracy, matters for iterative generation.

\paragraph{Preservation bias resolution.}
\label{sec:bias_resolution}
Figure~\ref{fig:calib_stress} (right) isolates the failure mode D3IM exposes: positions where the model's top-confidence self-commit is wrong. Original preserves the wrong commit 70.1\% of the time and recovers the ground-truth token only 9.2\% of the time. \ours{} reverses this behavior, reducing wrong-token preservation to 17.0\% and recovering the ground-truth token 50.2\% of the time. This is consistent with the ranking-calibration story: \ours{} trains the model to assign lower confidence to its own erroneous predictions.

\subsection{Main Results: D3IM Requires Training-Inference Alignment}

\begin{table*}[t]
\centering
\caption{Main results with full step scaling. Strategy denotes the inference rule: ``std unmask'' is monotonic low-confidence remasking with committed tokens frozen, while ``D3IM'' is clean-slate top-$K$ remasking with token revision enabled. GSM8K: 1319 problems, 8-shot CoT. MATH-500: 500 problems, sympy-verified. HumanEval: 164 problems, pass@1. MBPP: 500 problems, pass@1 (lm-evaluation-harness 3-shot). \ours{}: LoRA $r{=}64$, 500 steps on FineWeb-Edu. Bold = best in column.}
\label{tab:main}
\footnotesize
\setlength{\tabcolsep}{3pt}
\begin{tabular}{ll ccc ccc ccc ccc}
\toprule
& & \multicolumn{3}{c}{\textbf{GSM8K}} & \multicolumn{3}{c}{\textbf{MATH-500}} & \multicolumn{3}{c}{\textbf{HumanEval}} & \multicolumn{3}{c}{\textbf{MBPP}} \\
\cmidrule(lr){3-5}\cmidrule(lr){6-8}\cmidrule(lr){9-11}\cmidrule(lr){12-14}
\textbf{Model} & \textbf{Strategy} & 64 & 128 & 256 & 64 & 128 & 256 & 64 & 128 & 256 & 64 & 128 & 256 \\
\midrule
Original LLaDA & std unmask & 55.3 & 58.4 & 59.5 & 18.8 & 21.6 & 22.4 & 14.0 & 26.8 & 32.9 & 20.4 & 29.4 & \best{39.2} \\
Original LLaDA & D3IM & 48.0 & 51.6 & 55.3 & 17.8 & 18.6 & 19.2 & 5.5 & 4.3 & 7.3 & 6.8 & 8.8 & 12.0 \\
\midrule
\ours{} & std unmask & 52.8 & 55.3 & 59.7 & 19.2 & 21.2 & 27.0 & 6.7 & 20.1 & 31.1 & 16.4 & 27.6 & 36.0 \\
\ours{} & D3IM & \best{68.3} & \best{71.9} & \best{73.8} & \best{23.6} & \best{27.6} & \best{30.0} & \best{29.3} & \best{30.5} & \best{34.8} & \best{30.8} & \best{36.8} & 36.6 \\
\bottomrule
\end{tabular}
\end{table*}

Our primary comparison is against standard LLaDA decoding (Original + std unmask), since this is the inference method used by the unmodified model. At 64 steps, \ours{}+D3IM improves GSM8K from 55.3\% to 68.3\%, MATH-500 from 18.8\% to 23.6\%, HumanEval from 14.0\% to 29.3\%, and MBPP from 20.4\% to 30.8\%. The Original+D3IM row answers a different question: whether committed-token revision works without training alignment.

Two observations confirm that \ours{}'s gains are sampler-matched rather than a generic training improvement. First, \ours{} under standard unmasking (\ours{} training without D3IM sampling) remains comparable to Original on math but is weaker on code (Table~\ref{tab:main}, ``std unmask'' rows), showing that the training does not uniformly improve monotonic decoding. Second, D3IM on the unmodified model (D3IM sampling without \ours{} training) is \emph{harmful} across all four benchmarks at every step count (e.g.\ GSM8K drops from 55.3\% to 48.0\% at 64 steps), and the gap does not close with more steps. The failure is not due to an insufficient step budget, but to unreliable confidence ranking under self-generated context: once D3IM exposes visible self-commits, the unadapted model cannot reliably decide which ones should survive.

For \ours{} training, D3IM sampling is \emph{beneficial} and the advantage grows with the step budget on math and reasoning tasks. At 64 steps, \ours{}+D3IM scores 68.3\% on GSM8K ($+$15.5 over \ours{} std); at 256 steps this rises to 73.8\%, a $+$18.5-point gap over Original+D3IM at the same budget. The same scaling pattern holds on MATH-500 (23.6\% $\to$ 30.0\%) and HumanEval (29.3\% $\to$ 34.8\%). MBPP is the main exception: \ours{}+D3IM improves substantially at 64 and 128 steps (30.8\%, 36.8\%) but saturates at 256 steps (36.6\%) and does not surpass Original standard unmasking (39.2\%). This suggests that clean-slate revision is most effective when additional steps help resolve earlier semantic errors, while long code completion may benefit from a softer or more length-aware revision policy.

\subsection{Trajectory Analysis}
\label{sec:trajectory}

\begin{table}[t]
\centering
\caption{D3IM correction examples on GSM8K. \textbf{Left}: Original model on problem \#29 (answer\,=\,23); 274 corrections total, the same positions oscillate between content and whitespace. \textbf{Right}: \ours{} model across several problems (all answered correctly); each correction targets a specific semantic error.}
\label{tab:case_study}
\footnotesize
\setlength{\tabcolsep}{3pt}
\begin{tabular}{@{}p{0.46\linewidth}p{0.46\linewidth}@{}}
\toprule
\textbf{Original} & \textbf{\ours{}} \\
\midrule
\emph{Content $\to$ whitespace oscillation:} & \emph{Unit corrections:} \\[2pt]
``the'' $\to$ \texttt{\textbackslash n} & ``hour'' $\to$ ``minute'' \\
``the'' $\to$ \texttt{\textbackslash n} & ``week'' $\to$ ``hours'' \\
``the'' $\to$ \texttt{\textbackslash n} & ``div'' $\to$ ``ounces'' \\[3pt]
\texttt{\textbackslash n} $\to$ ``the'' (reversal) & \emph{Numerical recalculation:} \\[2pt]
``the'' $\to$ \texttt{\textbackslash n} (re-erased) & ``1''$\to$``2'', ``5''$\to$``0'' \\
\emph{Number erasure:} & ``=''$\to$``*'' (operator fix) \\[2pt]
``.'' $\to$ \texttt{\textbackslash n} & \emph{Concept / entity fixes:} \\[2pt]
``0'' $\to$ \texttt{\textbackslash n} & ``distance''$\to$``time'' \\
``0'' $\to$ \texttt{\textbackslash n} & ``groups''$\to$``boxes'' \\
\bottomrule
\end{tabular}
\end{table}

Table~\ref{tab:case_study} illustrates the qualitative difference. Without \ours{} training, D3IM corrections on a single GSM8K problem are dominated by content-to-whitespace oscillation: the same position flips between a content token and a newline across consecutive steps without settling. These oscillatory edits consume the step budget without improving the answer. With \ours{}, each correction instead targets a specific semantic error, such as wrong units (``hour''$\to$``minute''), incorrect digits, or mismatched operators, and positions are typically revised only once.

\subsection{Ablations and Additional Analysis}
\label{sec:ablations}

\paragraph{Design choice ablation.}
We report leave-one-out ablations from the full \ours{} recipe in Table~\ref{tab:ablation}. Each row removes one design choice while keeping the LoRA rank, data, training steps, and D3IM inference fixed. The \emph{MDM-only} row disables the entire self-conditioning branch and serves as a matched-budget continued-training baseline rather than a strict leave-one-out variant. The two largest contributors are focused loss ($-10.8$) and confidence-based commit ($-3.2$): supervising all positions reduces the relative weight of the error-correction gradient, while random commit removes the model's ability to lock in high-confidence tokens. Notably, MDM-only continued training (46.2\%) performs even worse than Original+D3IM without any training (48.0\%), suggesting that continued training on the standard MDM objective alone shifts the model away from D3IM's decision boundary without improving its revision ability.

\begin{table}[t]
\centering
\caption{Leave-one-out ablation of \ours{} design choices. All variants use LLaDA-8B, LoRA $r{=}64$, 500 steps, FineWeb-Edu, D3IM-64. Drop is relative to full \ours{}. MDM-only is a matched-budget baseline, not a strict leave-one-out row.}
\label{tab:ablation}
\footnotesize
\begin{tabular}{lcc}
\toprule
\textbf{Variant} & \textbf{GSM8K} & \textbf{Drop} \\
\midrule
Original + D3IM, no training & 48.0 & $-20.3$ \\
MDM-only continued training & 46.2 & $-22.1$ \\
\midrule
Full \ours{} & \best{68.3} & -- \\
\midrule
$-$ Focused loss (all-position CE) & 57.5 & $-10.8$ \\
$-$ Confidence commit (random) & 65.1 & $-3.2$ \\
\bottomrule
\end{tabular}
\end{table}

\paragraph{D3IM channel ablation.}
Table~\ref{tab:channel_ablation} isolates D3IM's two correction channels at 64 steps on bounded subsets (GSM8K-200; MBPP-50). On the \ours{} checkpoint, both channels contribute: disabling either token-to-token revision or token-to-mask demotion reduces accuracy, and disabling both recovers standard unmasking performance. On the Original checkpoint, the pattern reverses for token-to-token revision: disabling it raises GSM8K from 47.0\% to 63.0\%, indicating that without \ours{} the model's direct token overwrites are destructive. After \ours{} training, the same channel becomes productive (full D3IM reaches 70.5\%), confirming that preservation bias, not sampler design, is the bottleneck.

\begin{table}[t]
\centering
\caption{D3IM channel ablation at 64 steps on GSM8K-200 and MBPP-50. ``No-t2t'' disables token-to-token overwrite; ``no-t2m'' disables token-to-mask demotion; ``masked-only'' disables both. Bold = best per model.}
\label{tab:channel_ablation}
\footnotesize
\setlength{\tabcolsep}{4pt}
\begin{tabular}{@{}l rr rr@{}}
\toprule
\multirow{2}{*}{\textbf{Sampler}} & \multicolumn{2}{c}{\textbf{Original}} & \multicolumn{2}{c}{\textbf{\ours{}}} \\
\cmidrule(lr){2-3}\cmidrule(lr){4-5}
& GSM & MBPP & GSM & MBPP \\
\midrule
std unmask            & 59.5 & 8.0  & 59.5 & 6.0 \\
D3IM masked-only      & 59.5 & 8.0  & 59.0 & 6.0 \\
D3IM no-t2t           & \best{63.0} & 4.0  & 66.0 & \best{10.0} \\
D3IM no-t2m           & 39.5 & 0.0  & 64.0 & 4.0 \\
D3IM (full)           & 47.0 & 2.0  & \best{70.5} & \best{10.0} \\
\bottomrule
\end{tabular}
\end{table}

\paragraph{Sampler-specificity.}
\label{sec:sampler_specificity}
\ours{} training is matched to D3IM's clean-slate top-$K$ decision rule. Table~\ref{tab:inference_baselines} applies three inference-only revokable samplers (ReMDM-conf, ReMDM-cap~\cite{remdm}, and Tolerator~\cite{tolerator}) on both Original and \ours{}-trained checkpoints. D3IM gains ${+}6$ to ${+}24$\,pp from \ours{} training across the three benchmarks, while ReMDM is consistently hurt ($-1$ to $-4$\,pp) and Tolerator is essentially neutral. This is expected: \ours{} specifically trains the model's \emph{token $\to$ token} revision ability by exposing it to self-generated visible errors and teaching it to directly replace wrong tokens with correct ones. D3IM exploits this ability through its global top-$K$ re-ranking. ReMDM, however, only uses the \emph{mask $\to$ token} pathway: it first demotes a token to [MASK] and then re-predicts, so the token-to-token capability that \ours{} improves is never exercised. Tolerator uses a verify-and-refine loop independent of the confidence ranking, so it is neither helped nor hurt; it already reaches 65.4\% on GSM8K without \ours{}, but \ours{}+D3IM surpasses it (68.3\%).

\begin{table}[t]
\centering
\caption{Inference-only revokable samplers vs.\ \ours{}+D3IM at $N{=}64$ steps. \ours{} training strongly improves D3IM but does not transfer to ReMDM or Tolerator.}
\label{tab:inference_baselines}
\footnotesize
\setlength{\tabcolsep}{3pt}
\begin{tabular}{@{}l cc cc cc@{}}
\toprule
& \multicolumn{2}{c}{\textbf{HE}} & \multicolumn{2}{c}{\textbf{GSM}} & \multicolumn{2}{c}{\textbf{MATH}} \\
\cmidrule(lr){2-3}\cmidrule(lr){4-5}\cmidrule(lr){6-7}
\textbf{Sampler} & Orig & \ours{} & Orig & \ours{} & Orig & \ours{} \\
\midrule
std unmask & 14.0 & 6.7  & 55.3 & 52.8 & 18.8 & 19.2 \\
D3IM &  5.5 & \best{29.3} & 48.0 & \best{68.3} & 17.8 & \best{23.6} \\
\midrule
ReMDM-conf     &  9.8 &  6.1 & 54.8 & 52.5 & 16.2 & 14.2 \\
ReMDM-cap      &  6.7 &  5.5 & 52.1 & 47.8 & 14.6 & 14.4 \\
Tolerator  & 21.3 & 21.3 & 65.4 & 65.7 & 19.6 & 17.2 \\
\bottomrule
\end{tabular}
\end{table}

Additional ablations covering temperature ($\tau$) and commit-rate schedule ($\rho$) are reported in Appendix~\ref{sec:appendix_additional_ablations}.

\section{Related Work}
\label{sec:related}

\paragraph{Training on self-generated states.}
Several recent methods reduce the train-inference gap by exposing diffusion language models to self-generated states. DMax~\cite{dmax} introduces On-Policy Uniform Training and pairs it with Soft Parallel Decoding in an interpolated embedding space. ProSeCo~\cite{proseco} trains a corrector from the model's own unmasking errors and adds corrective refinement steps during generation. LLaDA2.1~\cite{llada21} jointly trains Mask-to-Token and Token-to-Token objectives, enabling threshold-based token editing at inference. These methods share with \ours{} the idea of training on self-generated predictions, but differ in interface and goal: DMax targets soft embedding-space decoding, ProSeCo adds explicit correction passes, and LLaDA2.1 relies on thresholded token editing. \ours{} training instead aligns the confidence ranking with D3IM sampling's clean-slate top-$K$ rule.

\paragraph{Inference-time revokable decoding.}
A complementary line of work makes committed tokens revisable at inference time without retraining the base model. ReMDM~\cite{remdm} derives a stochastic remasking process for absorbing-state MDMs; WINO~\cite{wino} uses draft-and-verify decoding; Tolerator~\cite{tolerator} interleaves fill-up with refinement; CoRe~\cite{core} probes context sensitivity to detect brittle tokens; and T2M~\cite{t2m} resets suspect tokens to [MASK] rather than directly overwriting them. These methods typically decide which existing tokens to revisit using remasking schedules, local detection rules, or verification passes. D3IM differs by constructing each intermediate state directly from current predictions through a global top-$K$ ranking. Our channel ablation further shows that the same revision channels are harmful without \ours{} but beneficial after sampler-matched training.

\paragraph{Confidence, calibration, and token-quality signals.}
Many remasking methods depend on confidence or quality estimates to decide which tokens should survive. PC-Sampler~\cite{pcsampler} calibrates position-dependent confidence biases, RCR~\cite{mdpo} adjusts remasking thresholds, and learned-score methods such as RemeDi~\cite{remedi} and PRISM~\cite{prism} introduce token-quality or confidence signals for refinement. These approaches are complementary to our motivation, but our results suggest that confidence improvements are sampler-specific: \ours{} aligns confidence with D3IM's clean-slate top-$K$ competition and does not automatically transfer to arbitrary revokable samplers. Additional training-side approaches~\cite{adaptive_order,gidd,backplay,d1} and remasking dynamics~\cite{corrective_dlm,lost_in_diffusion,starr,early_decisions,diffucoder,idlm,soft_masked} are discussed in Appendix~\ref{sec:appendix_remasking_dynamics}.

\section{Discussion}
\label{sec:discussion}

\paragraph{Two revision pathways.}
Since D3IM sets $c_t = 0$ and re-ranks all positions by confidence at each step, a previously committed token can either be directly overwritten by a higher-confidence prediction (\emph{token $\to$ token}) or lose the top-$K$ competition and be reset to [MASK] (\emph{token $\to$ mask}). These two pathways carry different signals: a direct replacement preserves the information that the position needs a \emph{similar} token (e.g.\ ``hour'' $\to$ ``minute''), while a demotion to [MASK] signals that the model lacks a reliable candidate and should re-predict from updated context. Retaining both gives D3IM more expressive revision than mask-only methods.

\paragraph{Exposure bias in bidirectional models.}
In bidirectional MDLMs, a wrong committed token is visible to \emph{all} positions, so subsequent generation may condition on the error and propagate it. This effect can be self-reinforcing: in autoregressive models, a committed token's confidence is fixed once generated, since only later positions attend to it; in bidirectional models, the wrong token also attends to subsequently generated context, and as surrounding tokens accommodate the error, the model's confidence in the wrong token may increase rather than decrease. Revision offers a second option: overwrite the wrong token instead of building around it. Our answer-first stress test (Appendix Table~\ref{tab:order_robustness}) is consistent with this hypothesis: \ours{}+D3IM achieves 68.0\% vs.\ 66.0\% reasoning-first, suggesting that revision mitigates error propagation even when answer tokens are likely committed with sparse context.

\section{Conclusion}
\label{sec:conclusion}

We showed that self-correcting inference in MDLMs requires aligning the sampler with the model's confidence behavior. D3IM re-evaluates every position at each step, but without \ours{} training the confidence ranking is unreliable and revision is harmful. \ours{} closes this gap through lightweight post-training, yielding gains specific to D3IM's token-to-token pathway that scale with the step budget. More broadly, our results suggest that the bottleneck for committed-token revision is the reliability of the confidence ranking under self-generated context, not the ability to revisit tokens itself.

Two directions remain open. First, D3IM's revision trajectory records which tokens were replaced or demoted at each step, providing step-level supervision analogous to process reward models. Second, recent RL approaches for diffusion language models collect rollouts under a fixed sampling strategy; D3IM's richer action space (token-to-token and token-to-mask at every step) may yield a higher RL ceiling by enabling revision policies unavailable under monotonic decoding.

\newpage
\section*{Limitations}

Token-level remasking corrects local errors such as wrong units, digits, or operators, but cannot restructure an entire reasoning process; when a problem requires a fundamentally different solution path, reasoning-level strategies such as best-of-$N$ or diffusion-based reasoning scaling~\cite{d1} are needed. All results are on LLaDA-8B; generalization to other MDLMs and larger scales remains to be verified. \ours{} training doubles per-step cost (two forward passes), though the total budget is small (500 LoRA steps, ${\sim}$20\,min on one A100). Hyperparameters ($\tau{=}1.5$, $\rho{=}0.3$) are fixed without systematic search; all numbers are single-run without error bars. We lack a theoretical account of why single-step self-conditioning generalizes to 64-step inference.

\bibliography{references}

\clearpage
\appendix

\section{Formal Details: Absorption, LLaDA, and D3IM}
\label{sec:appendix_formal_background}

This section provides the formal derivation underlying the intuition in \S\ref{sec:background}: LLaDA's training loss does not force the model to copy visible tokens, which is why revision-based inference like D3IM is possible. Readers familiar with discrete flow matching may skip to \S\ref{sec:appendix_preservation_bias}.

\paragraph{Notation.}
$\delta_y(\cdot)$ denotes the Dirac measure: $\delta_y(x)=1$ if $x=y$ and $0$ otherwise. $p_{0|t}(\cdot \mid X_t)$ denotes the posterior distribution of the clean token given the current state at time $t$.

\subsection{Standard masked diffusion models}
\label{sec:dfm_background}\label{sec:remdm_background}

Standard masked diffusion models (e.g., Discrete Flow Matching~\citep{dfm}, ReMDM~\citep{remdm,mdlm}) define a probability path using a coordinate-wise mixture kernel:
\begin{equation}
  p_t(x_t^i \mid x_0, x_1)
  =
  \kappa_t\,\delta_{x_0^i}(x_t^i)
  +
  (1 - \kappa_t)\,\delta_{x_1^i}(x_t^i),
\end{equation}
where $\kappa_t \in [0,1]$ satisfies $\kappa_0 = 1$ (clean) and $\kappa_1 = 0$ (fully masked). Since the endpoint $x_1$ is fully masked, $p_{1 \mid t}(x_1^i \mid x_t) = \delta_m(x_1^i)$, which implies:
\begin{equation}
  p_{0 \mid t}(X_0^i \mid X_t, X_t^i \neq m) = \delta_{X_t^i}.
\end{equation}
This is the \emph{absorbing posterior property}: once a token is unmasked, subsequent reverse steps cannot change it. The training objective matches this by supervising \emph{all} positions:
\begin{equation}
\label{eq:absorbing_mdm_ce}
\mathcal{L}_{\mathrm{abs}}(\theta)
=
\mathbb{E}_{x_0,\,t,\,x_t,\,i}
\Big[
- \log p_\theta^i(x_0^i \mid x_t, t)
\Big].
\end{equation}

\subsection{LLaDA's training objective}
\label{sec:llada_objective}

LLaDA uses a masked-token cross-entropy that supervises only masked positions:
\begin{equation}
\label{eq:llada_loss}
\mathcal{L}_{\mathrm{llada}}(\theta)
= \mathbb{E}\!\biggl[
\sum_{i=1}^N \mathbf{1}[x_t^i {=} m]
\bigl({-}\log \pi_\theta^i(x_0^i \mid x_t)\bigr)
\biggr].
\end{equation}
Because the loss imposes no constraint at unmasked positions, the model's prediction $\pi_\theta^i(\cdot \mid x_t)$ is not trained to satisfy the absorbing identity $\pi_\theta^i(\cdot \mid x_t)=\delta_{x_t^i}$ when $x_t^i \neq m$. The objective is better understood as a masked pseudo-likelihood that encourages local conditional consistency without enforcing global path consistency.

\subsection{Preservation bias despite non-absorbing predictions}
\label{sec:appendix_preservation_bias}

The absence of an absorbing constraint means LLaDA can in principle revise committed tokens---an opportunity that D3IM exploits. However, this corrective capacity is limited in practice: we term this \emph{preservation bias}. The bias does not arise from the training objective enforcing token preservation, but from the model never having seen its own erroneous predictions as context during training. See \S\ref{sec:bias_resolution} for quantitative measurements and \S\ref{sec:scope_training} for how \ours{} closes this gap.

\subsection{D3IM as a corrector sampler}
\label{sec:appendix_theory}

D3IM can be understood within the general corrector sampling framework for discrete diffusion~\cite{dfm}. Using Euler discretization, the corrector update at each step takes the form:
\begin{equation}
\label{eq:corrector}
X_{t+h}^i \sim a_t\, p_{1|t}(\cdot \mid X_t) + b_t\, p_{0|t}(\cdot \mid X_t) + c_t\, \delta_{X_t^i}(\cdot),
\end{equation}
subject to a probability constraint $a_t + b_t + c_t = 1$ and a noise-conservation constraint $b_t + c_t(1 - \kappa_t) = 1 - \kappa_{t+h}$, where $\kappa_t$ is the masking schedule ($\kappa_0 = 1$ for fully unmasked, $\kappa_1 = 0$ for fully masked). Here $p_{1|t}$ is the data posterior (predicted clean token) and $p_{0|t}$ is the noise posterior (which collapses to $\delta_m$ in masked diffusion).

Standard absorbing samplers set $b_t = 0$ (no remasking) and assign full weight to $c_t$ (preserve current token) and $a_t$ (unmask via data posterior), yielding monotonic unmasking where committed tokens are frozen.

Recent remasking methods introduce $b_t > 0$, allowing tokens to be demoted back to [MASK]. ReMDM~\cite{remdm} remasks low-confidence positions; WINO~\cite{wino} drafts a full sequence then selectively revokes tokens; CoRe~\cite{core} and T2M~\cite{t2m} apply context-aware or token-to-mask refinement rules. Despite these differences, all such methods retain $c_t > 0$: committed tokens carry preservation weight, and any revision must route through the mask state (token$\to$[MASK]$\to$token across two or more steps). The revision pathway is therefore indirect---no single step can replace one visible token with a different visible token.

D3IM instead operates with $c_t = 0$: no weight is placed on preserving the current token. Every position is re-evaluated through $p_{1|t}$ at each step, and the top-$K$ most confident predictions survive while the rest are reset to [MASK] (effectively resampled through $p_{0|t} = \delta_m$). This opens a direct token$\to$token channel absent from all absorbing and indirect-remasking samplers, giving D3IM higher revision capacity within the same framework, at the cost of requiring reliable confidence rankings under self-generated context---the gap that \ours{} training closes.

\section{Algorithm Details}
\label{sec:algo_details}

\begin{algorithm}[h]
\caption{D3IM Inference (clean-slate remasking)}
\label{alg:D3IM}
\KwInput{Model $f_\theta$, prompt $p$, length $L$, steps $T$}
$x \gets \lbrack\text{MASK}\rbrack^L$\tcp*{fully masked}
\For{$t \leftarrow 1$ \KwTo $T$}{
  $z \gets f_\theta(\lbrack p;\, x\rbrack)$\tcp*{predict ALL positions}
  $x_0 \gets \arg\max(z)$\tcp*{best prediction per position}
  $c \gets \max(\text{softmax}(z))$\tcp*{confidence per position}
  $K \gets \lfloor L \cdot t / T \rfloor$\tcp*{linear schedule}
  $\textit{top} \gets \text{top-}K\text{ positions by } c$\;
  $x \gets \lbrack\text{MASK}\rbrack^L$\tcp*{reset everything}
  $x\lbrack\textit{top}\rbrack \gets x_0\lbrack\textit{top}\rbrack$\tcp*{only top-$K$ survive}
}
\Return{$x$}
\end{algorithm}

\begin{algorithm}[h]
\caption{\ours{} (Self-Conditioned On Prediction Errors) Training (one step)}
\label{alg:scope}
\KwInput{Model $f_\theta$, tokens $x$, commit rate $\rho$, temperature $\tau$}
$r \sim \text{Beta}(2,2)$; mask $r$\% of positions $\to x_{\text{masked}}$\;
$\hat{z} \gets f_\theta(x_{\text{masked}})$ \tcp*{no gradient}
$\hat{x} \sim \text{Cat}(\text{softmax}(\hat{z}/\tau))$ \tcp*{temperature sample}
$c \gets \text{confidence}(\hat{z})$ \tcp*{temperature $=1$ logits}
$C \gets \text{top-}\rho\text{\% of masked positions by } c$\;
$x_{\text{train}} \gets x$; \quad $x_{\text{train}}\lbrack C\rbrack \gets \hat{x}\lbrack C\rbrack$\;
$M \gets \text{masked} \setminus C$ \tcp*{remaining masks}
$x_{\text{train}}\lbrack M\rbrack \gets \lbrack\text{MASK}\rbrack$\;
$z \gets f_\theta(x_{\text{train}})$ \tcp*{with gradient}
$\mathcal{L} \gets \text{mean}_{i \in C \cup M}\;\text{CE}(z_i, x_i^*)$\;
\Return{$\mathcal{L}$}
\end{algorithm}

This section documents the implementation details that affect numerical reproduction.

\subsection{D3IM Inference (semi-AR with last-step accept)}
\label{sec:appendix_d3im}

In addition to the clean-slate top-$K$ rule of Algorithm~\ref{alg:D3IM}, our reference implementation uses three details that we keep consistent across all D3IM evaluations.

\paragraph{Semi-autoregressive blocks.} The generation length $L$ is divided into $L/B$ blocks of $B$ tokens each, processed left-to-right. At block $k$, the schedule budget $K(t)$ is applied only within the current block: $K_k(t) = \lfloor B \cdot t / T \rfloor$ for $t \in \{1,\ldots,T\}$. Positions in earlier blocks ($<k$) are frozen and not re-evaluated; positions in later blocks ($>k$) remain $[\text{MASK}]$ until their block is reached. This matches the LLaDA reference sampler. With $B = L$ (single block, our default for short generations like GSM8K), the schedule reduces to the global form in Algorithm~\ref{alg:D3IM}.

\paragraph{Token-to-token (t2t) clean slate.} For non-D3IM samplers, an unmasked committed token is preserved when the model prediction $\hat{x}_i$ disagrees: $x_i \gets \mathbb{1}[i \in \mathcal{M}_t] \, \hat{x}_i + \mathbb{1}[i \notin \mathcal{M}_t] \, x_i$. D3IM omits this override: $x_0 = \arg\max(z)$ for \emph{every} position, so a high-confidence revision can replace a previously committed token. Without this t2t channel, the sampler can only demote and re-fill, not directly correct.

\paragraph{EOS / EOT confidence suppression.} LLaDA's vocabulary has explicit end-of-sequence tokens (ids 126081 and 126348). Without intervention, these tokens can receive high confidence and occupy survival slots early in the trajectory, destabilizing length-dependent metrics. During non-final denoising steps, if the top-1 candidate at a position is EOS/EOT, we set that position's confidence to $-\infty$ before the top-$K$ survival decision. This prevents EOS/EOT candidates from surviving through confidence ranking. We do not set their confidence to $+\infty$, and we do not use a separate confidence-preservation rule for EOS/EOT. At the final step, the block is accepted without another top-$K$ ranking.

\paragraph{Final-step accept.} At step $t = T$, $K(T) = L$ and the entire block is committed regardless of confidence (no positions remain for the next iteration). This guarantees the algorithm terminates with a fully-decoded sequence.

\subsection{EOS / EOT Policy Ablation}
\label{sec:eos_policy_ablation}

D3IM lets all tokens compete for top-$K$ at every step, so EOS/EOT tokens can win early and cause premature termination. We compare four policies (Table~\ref{tab:eos_policy}):
\textbf{(1)~No suppression}: EOS competes normally; 20\% of generations place EOS before position 128.
\textbf{(2)~Confidence suppression} (chosen): the model may still predict EOS, but during non-final steps its confidence is set to $-\infty$ so it cannot survive top-$K$ ranking. EOS only appears at the final step when all positions are accepted.
\textbf{(3)~Logit suppression, non-final}: EOS logits are zeroed before softmax at non-final steps, preventing EOS from being predicted at all. This is more aggressive: early EOS nearly vanishes, but 22.9\% of generations never produce EOS (truncated), and HumanEval drops to 14.0\%.
\textbf{(4)~Logit suppression, all steps}: EOS logits are zeroed at every step including the final one. All generations are truncated.
Confidence suppression achieves the best accuracy (GSM8K 70.5, HumanEval 46.0) by delaying EOS without removing it from the model's output distribution.

\begin{table}[t]
\centering
\caption{EOS/EOT policy ablation (\ours{}+D3IM, 64 steps). Task accuracy on bounded subsets (GSM8K 200, HumanEval 50, MBPP-smoke 100). First EOS pos.\ and Early EOS\,$<$128 are pooled over 350 generations; Trunc.\ = fraction with no EOS at all.}
\label{tab:eos_policy}
\footnotesize
\setlength{\tabcolsep}{2pt}
\begin{tabular}{@{}lccccc@{}}
\toprule
\textbf{Policy} & \textbf{GSM} & \textbf{HE} & \textbf{1st EOS} & \textbf{Early} & \textbf{Trunc.} \\
& & & \textbf{pos.} & \textbf{$<$128} & \\
\midrule
No suppress. & 64.5 & 36.0 & 208.8 & 20.0 & 0.0 \\
Conf.\ (chosen) & \best{70.5} & \best{46.0} & 233.2 & 8.0 & 0.0 \\
Logit, non-final & 65.0 & 14.0 & 254.0 & 0.3 & 22.9 \\
Logit, all steps & 64.0 & 14.0 & 256.0 & 0.0 & 100.0 \\
\bottomrule
\end{tabular}
\end{table}

\subsection{\ours{} Training: Implementation Details}
\label{sec:appendix_scope}

Algorithm~\ref{alg:scope} gives the simplified one-step procedure. The full implementation adds the following details for batched training with prompts. Hyperparameter defaults are $\rho = 0.3$, $\tau = 1.5$, $p_{\text{self-cond}} = 0.5$.

\paragraph{Prompt-mask handling.} Prompt positions $\mathcal{P} = \{i : i < \ell_p\}$ are never masked, so the self-prediction and the training forward pass both condition on the full prompt context. Only response positions participate in the mask--commit--retrain cycle.

\paragraph{Mask-rate clamping.} Per-sample mask rates $r_b \sim \text{Beta}(2,2)$ are clamped to $[0.1, 0.9]$ to avoid trivially easy (nearly unmasked) or trivially hard (nearly fully masked) training examples.

\paragraph{Per-sample top-$K$ commit.} Because each sample $b$ has a different mask count $|\mathcal{M}_b|$, we compute $K_b = \max(1, \lfloor \rho \cdot |\mathcal{M}_b| \rfloor)$ independently, then take the union $C = \bigcup_b C_b$ as the committed set.

\paragraph{Standard-MDM mixing.} With probability $1 - p_{\text{self-cond}}$, we skip the self-conditioning branch entirely and fall back to the standard MDM training objective: mask random positions and compute the CE loss only on masked positions. This preserves the model's ordinary mask-filling ability and prevents catastrophic forgetting of the base distribution.

\paragraph{Temperature branching.} When $\tau > 0$, self-predictions are drawn from $\text{Categorical}(\text{softmax}(z / \tau))$; when $\tau = 0$, we fall back to $\arg\max$. Confidence $c_i$ is always computed from the $\tau{=}1$ logits regardless of the sampling temperature, so that the commit ranking reflects calibrated probabilities.

\paragraph{Why three design choices in one recipe.} Table~\ref{tab:ablation} ablates the three departures from a vanilla self-conditioning recipe: (i)~predicting from a \emph{masked} context rather than a clean one (so the self-prediction sees the same masked context the inference loop would), (ii)~committing the top-$\rho$ fraction by confidence rather than replacing all masked positions, and (iii)~focusing the CE loss on $C \cup M$ rather than supervising all positions. Beta(2,2) mask-rate sampling and temperature sampling are sub-design choices: Beta(2,2) keeps the mask rate near 0.5 (avoiding trivially easy or trivially hard samples), and $\tau{=}1.5$ pushes the empirical error rate from ${\sim}5\%$ (argmax) to ${\sim}20$--$45\%$, better matching the error rate D3IM encounters at inference (Appendix Table~\ref{tab:temp_ablation}).

\paragraph{Monitoring metrics.} During training we track two split losses on the committed set $C$: $\mathcal{L}_{\text{wrong}}$ over the subset $C_w = \{i \in C : \hat{x}_i \neq x_i\}$ measures error correction, and $\mathcal{L}_{\text{right}}$ over $C_r = \{i \in C : \hat{x}_i = x_i\}$ measures preservation of correct commits. A healthy run shows $\mathcal{L}_{\text{wrong}}$ decreasing (model learns to correct) while $\mathcal{L}_{\text{right}}$ stays low (preservation is easy because the gold token was committed). On 50\% mask rate, training begins with $\mathcal{L}_{\text{wrong}} \approx 6.3$ and $\mathcal{L}_{\text{right}} \approx 0.8$; both decrease over 500 steps to $\mathcal{L}_{\text{wrong}} \approx 1.5$ and $\mathcal{L}_{\text{right}} \approx 0.05$.

\section{Additional Ablations}
\label{sec:appendix_additional_ablations}

\subsection{Temperature}
\label{sec:appendix_temp}

\begin{table}[t]
\centering
\caption{Temperature ablation for \ours{}'s self-prediction step (GSM8K, 64 steps, LoRA $r{=}64$). Higher sampling temperature improves standard unmasking and modestly improves D3IM.}
\label{tab:temp_ablation}
\footnotesize
\setlength{\tabcolsep}{6pt}
\begin{tabular}{c ccc}
\toprule
\textbf{$\tau$} & \textbf{std unmask} & \textbf{D3IM} & \textbf{D3IM $-$ std} \\
\midrule
0 (argmax)   & 42.2 & 66.0 & $+$23.8 \\
1.0          & 41.7 & 66.8 & $+$25.1 \\
1.5 (paper)  & 52.8 & 68.3 & $+$15.5 \\
2.0          & \best{55.2} & \best{68.7} & $+$13.5 \\
\bottomrule
\end{tabular}
\end{table}

Sweeping $\tau \in \{0, 1, 1.5, 2\}$ on the full GSM8K set (Table~\ref{tab:temp_ablation}) reveals two effects: higher $\tau$ lifts std-unmasking accuracy by 13 points (42.2\% $\to$ 55.2\%), but D3IM attenuates this effect (only $+$2.7), because D3IM's clean-slate rule can re-resolve mistakes from poor training. We use $\tau{=}1.5$; $\tau{=}2.0$ is marginally stronger.

\subsection{Commit-Rate Schedule}
\label{sec:appendix_commit_rate}

We tested replacing the fixed commit rate $\rho{=}0.3$ with three scheduled variants: uniform sampling $\rho \sim U(0.1,0.8)$, a curriculum from 0.1 to 0.8, and a narrow perturbation $\rho \sim U(0.2,0.4)$. None consistently improves over the fixed rate across all benchmarks: each schedule trades a small gain on one task for a regression on another. We therefore keep $\rho{=}0.3$ as the default.

\subsection{D3IM Channel Ablation: Extended Analysis}
\label{sec:appendix_channel_ablation}

Table~\ref{tab:channel_ablation} in the main text reports D3IM channel ablations on the core five sampler variants. Table~\ref{tab:channel_ablation_full} extends this comparison with inference-only revokable baselines evaluated on the same GSM8K-200 and MBPP-50 subsets at 64 steps. The additional rows confirm that \ours{} training specifically improves D3IM: WINO (draft-only, without its second verification pass) is unchanged (57.0\% $\to$ 57.0\%), ReMDM-conf gains only 1 point, and ReMDM-cap is hurt. Tolerator reaches 70.0\% on \ours{}, nearly matching full D3IM (70.5\%), but relies on a more complex two-phase inference loop and does not benefit from the channel decomposition that makes D3IM's gains interpretable.

\begin{table}[t]
\centering
\caption{Extended sampler comparison including inference-only revokable baselines, at 64 steps on GSM8K-200 and MBPP-50. Bold = best per model column. D3IM channel ablation rows are repeated from Table~\ref{tab:channel_ablation} for reference. $^\dagger$WINO without the second verification pass (draft-only).}
\label{tab:channel_ablation_full}
\footnotesize
\setlength{\tabcolsep}{4pt}
\begin{tabular}{@{}l rr rr@{}}
\toprule
\multirow{2}{*}{\textbf{Sampler}} & \multicolumn{2}{c}{\textbf{Original}} & \multicolumn{2}{c}{\textbf{\ours{}}} \\
\cmidrule(lr){2-3}\cmidrule(lr){4-5}
& GSM & MBPP & GSM & MBPP \\
\midrule
std unmask            & 59.5 & 8.0  & 59.5 & 6.0 \\
D3IM masked-only      & 59.5 & 8.0  & 59.0 & 6.0 \\
D3IM no-t2t           & \best{63.0} & 4.0  & 66.0 & \best{10.0} \\
D3IM no-t2m           & 39.5 & 0.0  & 64.0 & 4.0 \\
D3IM (full)           & 47.0 & 2.0  & \best{70.5} & \best{10.0} \\
\midrule
ReMDM-conf            & 59.0 & 4.0  & 60.0 & 6.0 \\
ReMDM-cap             & 54.0 & 4.0  & 47.5 & 2.0 \\
WINO$^\dagger$        & 57.0 & 8.0  & 57.0 & 8.0 \\
Tolerator             & \best{67.0} & 8.0  & 70.0 & 4.0 \\
\bottomrule
\end{tabular}
\end{table}

\subsection{Why Not Standard Calibration?}
\label{sec:appendix_calibration}

Post-hoc methods such as temperature scaling or Platt scaling rescale logits on a fixed validation distribution, but D3IM fails under a shifted input distribution containing the model's own wrong commits. Label smoothing similarly changes the target distribution, not the self-generated context. \ours{} keeps the ground-truth target and changes the training input, directly exposing the model to the contexts D3IM ranks at inference.

A natural concern is that \ours{} simulates only one D3IM step during training, yet inference runs $T{=}64$ steps whose context evolves iteratively. Our trajectory analysis (\S\ref{sec:trajectory}) suggests that single-step training is sufficient in this setting: the correction pattern persists across the full 64-step trajectory. One plausible explanation is that the relevant ranking behavior is local---deciding whether a visible token fits its surrounding context---and generalizes across progressive stages of D3IM inference.

\subsection{Comparison with XDLM Training}
\label{sec:appendix_xdlm}

XDLM~\cite{adaptive_order} introduces a stationary mixed-noise corruption process that unifies masked and uniform-noise diffusion language modeling. Its native sampler performs confidence-gated local token replacements, providing an explicit clean-token replacement path during inference. Unlike D3IM's global clean-slate survival rule, XDLM's decoder replaces a token only when the model's new prediction exceeds the current token's confidence, without demoting all positions to [MASK].

To test whether XDLM's training can substitute for \ours{}, we pair the publicly available LLaDA-XDLM checkpoint with both its native low-confidence sampler and D3IM (Table~\ref{tab:xdlm}).

\begin{table}[t]
\centering
\caption{XDLM vs.\ \ours{} on GSM8K (full 1319, 64 steps).}
\label{tab:xdlm}
\footnotesize
\begin{tabular}{@{}l l r@{}}
\toprule
\textbf{Model} & \textbf{Sampler} & \textbf{GSM8K} \\
\midrule
XDLM          & low-confidence & 50.8 \\
XDLM          & D3IM           & 60.3 \\
\ours{} (r64) & D3IM           & \best{68.3} \\
\bottomrule
\end{tabular}
\end{table}

Switching XDLM from its native sampler to D3IM improves GSM8K by +9.5 points (50.8\% $\to$ 60.3\%), indicating partial compatibility: D3IM's global remasking benefits even a model not trained for it. However, XDLM+D3IM (60.3\%) remains below \ours{}+D3IM (68.3\%). This supports our claim that D3IM benefits most from training that matches its hard top-$K$ survival dynamics, rather than from a generic mixed-noise objective.

\subsection{Exposure Bias Stress Test}
\label{sec:appendix_order_robustness}

As discussed in \S\ref{sec:discussion}, a wrong committed token in bidirectional models is visible to all positions and may propagate rather than get corrected. We stress-test this with an answer-first prompt format where the numerical answer precedes the chain-of-thought derivation. Because D3IM commits tokens by confidence rather than position, the answer is not guaranteed to appear first, but placing it before the reasoning increases the chance that answer tokens are committed while surrounding context is still sparse. We compare against the standard reasoning-first (CoT-first) format on 200 GSM8K examples at 64 steps.

\begin{table}[h]
\centering
\caption{Exposure bias stress test on GSM8K-200 (64 steps). CoT = reasoning-first, Ans = answer committed before reasoning.}
\label{tab:order_robustness}
\footnotesize
\begin{tabular}{@{}l l rr@{}}
\toprule
\textbf{Model} & \textbf{Sampler} & \textbf{CoT-first} & \textbf{Ans-first} \\
\midrule
Original & std unmask & 55.5 & 56.0 \\
Original & D3IM           & 46.0 & 55.0 \\
\ours{} (r64) & std unmask & 53.5 & 54.0 \\
\ours{} (r64) & D3IM           & \best{66.0} & \best{68.0} \\
\bottomrule
\end{tabular}
\end{table}

\ours{}+D3IM is the strongest configuration under both orderings, with only a 2-point gap between formats. In contrast, Original+D3IM shows a 9-point drop on CoT-first (46.0\%) compared to answer-first (55.0\%), suggesting that without \ours{} wrong early commits propagate through surrounding context. With \ours{}, D3IM can revise committed tokens as context develops, substantially reducing sensitivity to output order---consistent with the exposure bias hypothesis in \S\ref{sec:discussion}.

\section{Additional Related Work: Remasking Dynamics}
\label{sec:appendix_remasking_dynamics}

Corrective DLM~\cite{corrective_dlm} finds that confidence signals lack the resolution for precise error localization; Lost in Diffusion~\cite{lost_in_diffusion} documents failure modes including premature termination. STaRR~\cite{starr} analyzes spatial-temporal confidence dynamics and designs adaptive remasking thresholds; ``Early Decisions Matter''~\cite{early_decisions} uncovers a proximity bias where initial unmasking positions shape the trajectory; DiffuCoder~\cite{diffucoder} analyzes token generation order in code tasks. I-DLM~\cite{idlm} identifies an introspective consistency gap---DLMs often disagree with their own generations---and proposes modified training and decoding to close it, a framing related to our calibration perspective. Soft-Masked DLM~\cite{soft_masked} addresses information loss at the representation level by blending mask and token embeddings. These works analyze confidence patterns, denoising order, and failure modes, but none examine the revision process itself: which tokens get semantically corrected vs.\ superficially reshuffled, and how revision frequency relates to calibration quality. Our trajectory analysis (\S\ref{sec:trajectory}) fills this gap.

\end{document}